\documentclass[preprint,12pt,authoryear]{elsarticle}

\usepackage{amssymb}
\usepackage[utf8]{inputenc} 
\usepackage[T1]{fontenc}    
\usepackage{hyperref}       
\usepackage{url}            
\usepackage{booktabs}       
\usepackage{amsfonts}       
\usepackage{nicefrac}       
\usepackage{microtype}      
\usepackage{xcolor}         
\usepackage[ruled]{algorithm2e}
\usepackage{algorithmic}
\usepackage{amsmath}
\usepackage{amsthm}
\usepackage{multirow}

\usepackage{lineno}

\journal{Journal of \LaTeX\ Templates}

\begin{document}

\begin{frontmatter}



\title{DPSNN: A Differentially Private Spiking Neural Network with Temporal Enhanced Pooling}

\author[1,4,6]{Jihang Wang}
\ead{wangjihang2021@ia.ac.cn}
\author[1,4,6]{DongCheng Zhao}
\ead{zhaodongcheng2016@ia.ac.cn}
\author[1,3]{Guobin Shen}
\ead{shenguobin2021@ia.ac.cn}
\author[1,4]{Qian Zhang\corref{mycorrespondingauthor}}
\ead{q.zhang@ia.ac.cn}
\author[1,2,3,4,5]{Yi Zeng\corref{mycorrespondingauthor}}
\ead{yi.zeng@ia.ac.cn}
\cortext[mycorrespondingauthor]{Corresponding author}

\address[1]{Brain-inspired Cognitive Intelligence Lab, Institute of Automation, Chinese Academy of Sciences, Beijing, China}
\address[2]{Center for Excellence in Brain Science and Intelligence Technology, Chinese Academy of Sciences, Shanghai, China}
\address[3]{School of Future Technology, University of Chinese Academy of Sciences, Beijing, China}
\address[4]{School of Artificial Intelligence, University of Chinese Academy of Sciences, Beijing, China}
\address[5]{State Key Laboratory of Multimodal Artificial Intelligence Systems, Institute of Automation, Chinese Academy of Sciences, Beijing, China}

\fntext[6]{These authors have contributed equally to this work and share first authorship}

\begin{abstract}
Privacy protection is a crucial issue in machine learning algorithms, and the current privacy protection is combined with traditional artificial neural networks based on real values. Spiking neural network (SNN), the new generation of artificial neural networks, plays a crucial role in many fields. Therefore, research on the privacy protection of SNN is urgently needed. This paper combines the differential privacy(DP) algorithm with SNN and proposes a differentially private spiking neural network (DPSNN). The SNN uses discrete spike sequences to transmit information, combined with the gradient noise introduced by DP so that SNN maintains strong privacy protection. 
At the same time, to make SNN maintain high performance while obtaining high privacy protection, we propose the temporal enhanced pooling (TEP) method. It fully integrates the temporal information of SNN into the spatial information transfer, which enables SNN to perform better information transfer. We conduct experiments on static and neuromorphic datasets, and the experimental results show that our algorithm still maintains high performance while providing strong privacy protection.
\end{abstract}




\end{frontmatter}

\section{Introduction}\label{intro}

Machine learning has been used in applications rich in privacy data such as advertisement 
recommendation~\citep{konapure2021video} and health care~\citep{shahid2019applications}. Shopping habit data assists an operator's recommendation algorithm in recommending needed products, and health data enable medical intelligence to better assist doctors in treating patients. However, the model acquires better predictions by learning from a large amount of data, which has access to sensitive information. 
Recent studies~\citep{fredrikson2015model,melis2019exploiting,wei2018know,salem2020updates} have shown that sensitive information can be extracted from the deep neural network through various privacy attacks. Privacy-preserving measures should be introduced in a machine learning system to mitigate the privacy threats~\citep{du2017implementing,bost2014machine,yang2019federated}.

\begin{figure}[h!]
    \begin{center}
    \includegraphics[width=0.9\linewidth]{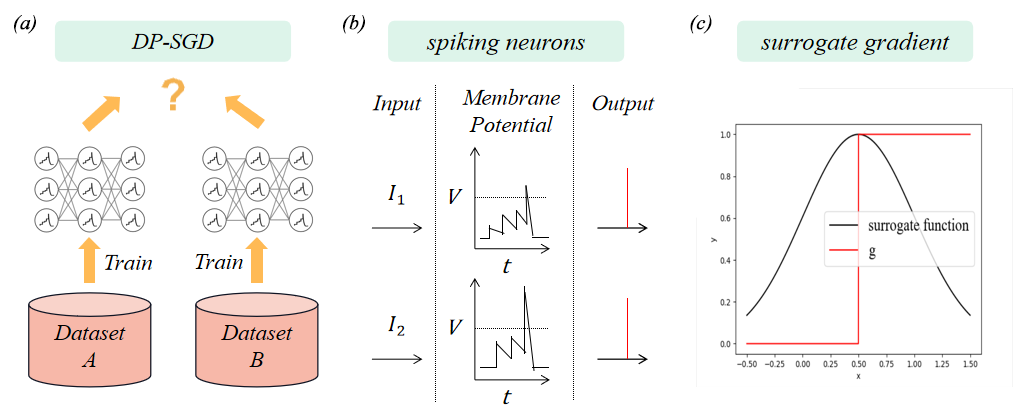}
    \end{center}
    \caption{This figure shows the privacy-preserving measures in the differentially private spiking neural network(DPSNN).}\label{introduction}
\end{figure}

Differential Privacy (DP) is widely used to protect privacy. DP adds the random mechanism to the data processing, making it difficult for observers to judge small changes in the dataset through the output results. If a machine learning model is differential private, every single data record in its training dataset contributes very little to the model's training as shown in Fig.~\ref{introduction}(a). Thus the privacy of the data is well protected. 
DP theory can measure the difficulty in distinguishing two datasets from the perspective 
of an attacker. $(\varepsilon ,\delta )$-DP define the privacy bound mathematically~\citep{dwork2016calibrating}, and the moments accountant
technique can analyse the $(\varepsilon ,\delta )$-DP 
in deep learning~\citep{abadi2016deep}. \citet{mironov2017renyi} use Renyi divergence to measure the distance between two probability distributions. They define Renyi differential privacy (RDP) based on Renyi divergence, and deduce the transformation from RDP to $(\varepsilon ,\delta )$-DP. However, this definition can only achieve a relatively loose privacy bound~\citep{bu2020deep}. \citep{dong2021gaussian} define DP through a hypothesis testing problem. They propose the 
definition of $f$-DP and a series of theories about gaussian differential privacy (GDP), which get a tighter bound of privacy estimation and is easier to handle the composition of private mechanisms. $\mu$-GDP
can be used to estimate the privacy bound, where $\mu $ means the difficulty in distinguishing the Gaussian distribution $N (0,1)$ from $N (\mu ,1)$. The smaller the parameter $\mu $, the better the privacy guarantees. 

The differentially private stochastic gradient descent (DP-SGD) algorithm~\citep{abadi2016deep} achieves DP on deep learning with random noise added to the clipping gradient. DP-SGD has achieved great success in many domains, such as medical imaging~\citep{ziller2021medical,li2019privacy}, Generative Adversarial Networks 
 (GANs)~\citep{hitaj2017deep,jordon2018pate,xie2018differentially}, and traffic flow estimation~\citep{cai2019differential}.

While the DP-SGD algorithm effectively stops known attacks in the neural network, the 
noise added in gradient descent will reduce the accuracy of the models. There have been 
many related works to improve the accuracy of DP-SGD.  \citep{yu2021gradient} proposes expected curvature
to improve the utility analysis of DP-SGD for convex optimization.
\citep{bu2020deep} uses $\mu $-GDP to estimate tighter privacy bounds of DP-SGD in deep learning model. 
\citep{papernot2021tempered} replaces the Rectified Linear Unit(ReLU) 
activation with the tempered sigmoid activation in the network, which can prevent 
the activations from exploding to improve the accuracy of 
differential private neural network trained by DP-SGD. \citet{tramer2020differentially} use handcrafted features in the network,
so that the network don't need to learn feature extraction layer by DP-SGD. They show that feature extraction method 
plays an important role in DP learning. 

Spiking neural network (SNN) also has some privacy-friendly characteristics. 
The non-real-valued information transmission greatly protects privacy. From Fig.~\ref{introduction}(b) we can see that, even though the input currents are different, they all reach the threshold and thus produce the same output. 
The training of SNN can be mainly divided into three categories: synaptic plasticity based~\citep{diehl2015unsupervised,hao2020biologically}, 
conversion based~\citep{li2021bsnn,han2020deep}, and backpropagation based~\citep{wu2018spatio}. The introduction of the surrogate gradient makes the backpropagation algorithm successfully applied to the training of SNN~\citep{neftci2019surrogate,wu2018spatio,jin2018hybrid,zhang2020temporal,shen2022backpropagation}. In addition to facilitating SNN's training, Fig.~\ref{introduction}(c) shows that the imprecise surrogate gradient makes it difficult for attackers to recover the users' information about the samples.

This study proposes a differentially private spiking neural network(DPSNN) that combines the characteristics of SNN and DP as well. 
As shown in Fig.~\ref{introduction}, the spiking neurons transmit information using spike sequences, the inexact derivative of the surrogate gradient, and the DP-SGD all provide a strong guarantee of the users' privacy.

DPSNN shows high accuracy with strong privacy guarantees. The contributions of our study can be summarized as follows:

\begin{itemize}
    \item To our best knowledge, this study is the first to combine DP with SNN to protect the privacy of SNN. 
    \item We propose a temporal enhanced pooling(TEP) method, which fully integrates the timing and spatial information of SNN, improving the trade-off between performance and privacy.
    \item We conduct experiments on static datasets MNIST~\citep{lecun1998mnist}, Fashion-MNIST~\citep{xiao2017fashion} and CIFAR10, as well as on the neuromorphic datasets N-MNIST~\citep{orchard2015converting} and CIFAR10-DVS~\citep{li2017cifar10}. The experimental results shows the superiority of TEP. 
\end{itemize}

\section{Preliminaries and background information}
In this section we introduce the definition of $(\varepsilon ,\delta )$-DP and RDP, the computational model of spiking neuron, 
the related work about privacy protection in machine learning and the existing algorithm with DP. 
\subsection{Differential Privacy Theory}
Two datasets are adjacent if there is only one different record between
each other. DP is aiming to make any pairs of datasets that differ in a single individual record difficult to distinguish. 
For the definition of $(\varepsilon ,\delta )$-DP, we refer to Dwork's previous work:

\newtheorem{definition}{Definition}
\begin{definition}[$(\varepsilon ,\delta )$-DP~\citep{dwork2016calibrating}]
    For any pair of adjacent datasets $D$, $D'$ and any event $E$, a randomized mechanism $M$ is $(\varepsilon ,\delta )$-DP if $\mathbb{P}(M(D)\in E)\leqslant
    e^{\varepsilon }\mathbb{P}(M(D')\in E) + \delta $.
\end{definition}

This study performs privacy analysis in the framework of RDP theory~\citep{mironov2017renyi}. RDP is based on Renyi divergence: 
$D_{\alpha }(P\| Q)\triangleq \frac{1}{\alpha -1} logE_{x\sim Q}(\frac{P(x)}{Q(x)} )^{\alpha }$.
Then we show the definition of $(\alpha, \varepsilon) $-RDP:

\begin{definition}[$(\alpha, \varepsilon) $-RDP~\citep{mironov2017renyi}]
    For any pair of adjacent datasets $D$, $D'$, a randomized mechanism $M$ is $(\alpha, \varepsilon) $-RDP 
    if $D_{\alpha }(M(D)\|M(D'))\leqslant \varepsilon$.
 \end{definition}

The Gaussian mechanism for a non-random algorithm $f$, which takes a dataset $D$ as input, is defined as $M(D) = f(D) + N(0, \sigma^2)$.
To analyse the RDP of the Gaussian mechanism, we need to calculate the global sensitivity of $f$.

\begin{definition}[Global Sensitivity~\citep{dwork2016calibrating}]
    Given an algorithm $f$, $D$, $D'$ are adjacent datasets, then the global sensitivity of $f$ is:
    $$GS_f=\max_{D,D'} \left\lVert f(D)-f(D')\right\rVert $$
 \end{definition}

The RDP privacy bound for the Gaussian mechanism can be calculated by the following theorem.
\newtheorem{theorem}{Theorem}

\begin{theorem}[The Privacy Bound of the Gaussian Mechanism~\citep{mironov2017renyi}]\label{gm}
    If the global sensitivity of $f$ is $R$, then Gaussian mechanism $M(D) = f(D) + R\cdot N(0, \sigma^2)$ 
    satisfies $(\alpha ,\alpha /(2\sigma ^2))$-RDP.
\end{theorem}

The numerical privacy bound for the composition of Gaussian mechanism can be calculated by the Opacus framework~\citep{opacus}. 

\subsection{Spiking Neuron Model}

The Leaky Integrate-And-Fire (LIF) model is commonly used to describe the neuronal dynamics of SNN~\citep{wu2018spatio,jin2018hybrid,zhang2020temporal}. The cell membrane is treated as a capacitor, and the dynamics of LIF neurons can be governed by a differential equation:
\begin{equation}\label{Lif0}
    \tau \frac{dV(t)}{dt} =-V(t)+R I(t)
\end{equation}
where $\tau $ is the time constant, $R$ is the membrane resistance, $V(t)$ is the membrane potential 
at time $t$. $I(t)=\sum_{j=1}^{l} w_{ij}o_j$ denotes the pre-synaptic input current. When the membrane potential 
reaches a certain threshold, the neurons fire a spike and return to the resting 
potential (we set the resting potential as zero in our model).

For better simulation, we convert the Eq.~\ref{Lif0} into a discrete form with the Euler method 
into Eq.\ref{lif}:
\begin{equation}\label{lif}
    V_{i}^{t+1,n}=\lambda V_{i}^{t,n}(1-o_{i}^{t,n})+(1-\lambda )I_{i}^{t,n}  
\end{equation}
Parameter $\lambda = 1-\frac{1}{\tau } $ describe the leak rate.
$o_{i}^{t,n}$ indicates whether the neuron $i$ in layer $n$ fires a spike at time $t$ ($0\leq t\leq T$, $T$ is the 
time steps). $V_{i}^{t,n}$ and $I_{i}^{t,n}$ mean the membrane potential and input current
of the $i$-neuron in layer $n$ at time $t$.

The neuron will fire a spike when the membrane potential exceeds the threshold $V_{th}$, which is shown in Eq.~\ref{eq3}. $g$ is the spiking activation function.
\begin{equation}\label{eq3}
    o_{i}^{t+1,n}=g(V_{i}^{t+1,n})=\begin{cases}
        0,\quad V_{i}^{t+1,n}<V_{th} \\
        1,\quad V_{i}^{t+1,n}\geq V_{th}\end{cases}
\end{equation}

The biggest limitation that restricts the use of BP in SNN is the non-differentiable 
spike activation function $g$. The surrogate gradient utilizes the inexact 
gradients near the threshold to enable the BPTT algorithm to be successfully used in the 
training of SNN. Herein, we apply a commonly used surrogate gradient 
function as shown in Eq.\ref{qgate}.

\begin{equation}\label{qgate}
    \frac{dg}{dV}= \begin{cases}
        -2 \left\lvert V-V_{th}\right\rvert + 1 ,\quad \left\lvert V-V_{th}\right\rvert<\frac{1}{2}  \\
        0,\quad \left\lvert V-V_{th}\right\rvert>\frac{1}{2}\end{cases}
\end{equation}
\subsection{Privacy-preserving Methods for Machine Learning}
The most straightforward approach to dealing with private data is sanitization, which filters out sensitive data from datasets. However, sanitization cannot guarantee that all private data will be removed when the amount of data is large. Researchers have developed many methods for privacy-preserving machine learning. Homomorphic encryption (HE) is then proposed. HE allows models to process encrypted data and return encrypted results to the clients, who can then decrypt them and get the results they want without data leakage. Many basic machine learning algorithms can be implemented with HE, including
linear regression, K-means clustering~\citep{du2017implementing}, naive bayes, decision tree, and support vector machine~\citep{bost2014machine}. HE can also be applied in parts of the Gaussian 
process regression algorithm to improve the privacy-preserving~\citep{fenner2020privacy}. In some cases, data from different enterprises have to be used to get a better model, but nobody wants to share their data with others. Federated learning can solve this problem. The main idea of federated learning is to build machine learning models based on datasets that are distributed across multiple devices while preventing data leakage~\citep{yang2019federated}.

\subsection{Algorithm with Differential Privacy}

DP mechanisms introduce randomness to the learning algorithm. The approaches to introducing randomness can be divided into three ways according to the place of noise added: output perturbation, objective perturbation, and gradient perturbation. The output perturbation adds noise in output 
parameters after the learning process~\citep{wu2017bolt}. The objective perturbation adds noise in the objective function and minimizes the perturbed objective~\citep{iyengar2019towards}. The gradient perturbation injects noise to the gradients of parameters 
in each parameter update~\citep{abadi2016deep}. 

Gradient perturbation can well release the noisy gradient at each iteration without damaging the privacy guarantee \citep{dwork2014algorithmic}. DP-SGD algorithm~\citep{abadi2016deep} uses gradient perturbation in the stochastic gradient descent and becomes a standard privacy-preserving method for machine learning.
Patient Privacy Preserving SGD (P3SGD) adds designed Gaussian noise to the update for both privacy protection and model 
regularization~\citep{P3SGD}.

Compared to the regularization methods such as dropout~\citep{srivastava2014dropout} 
and weight decay~\citep{krogh1991simple},
DP has been proven to be a more effective method to preserve privacy~\citep{carlini2019secret}.
This is because regularization methods can only prevent overfitting, but the unintended 
memorization of training data will still occur even the network is not overfitting. DP has been widely used in many machine learning algorithms. \citep{li2020privacy} proposes a
differential private gradient boosting decision tree, which has a strong guarantee of DP while the accuracy loss is less than the previous algorithm. Differential
private empirical risk minimization minimizes the empirical risk while guaranteeing 
the output of the learning algorithm differentially private with respect 
to the training data~\citep{bassily2014private}. 
\citep{phan2016differential} proposes deep private auto-encoders,
where DP is introduced by perturbing the cross-entropy errors. The adaptive Laplace mechanism preserves DP in deep learning by adding adaptive Laplace noise into affine transformation and loss function~\citep{phan2017adaptive}, where the privacy bound is independent of the number of training epochs. 

As the third-generation artificial neural network~\citep{maass1997networks}, the SNN is gradually applied in many domains. Also, the characteristics of SNN are more conductive to build a network with high privacy protection as we mentioned in section~\ref{intro}. This paper is the first attempt to effectively combine SNN and DP to build a network with high privacy guarantee and less accuracy loss. 

\section{Methods}
    \subsection{Differentially Private Spiking Neural Network}

\begin{figure}
        \begin{center}
        \includegraphics[width=0.9\linewidth]{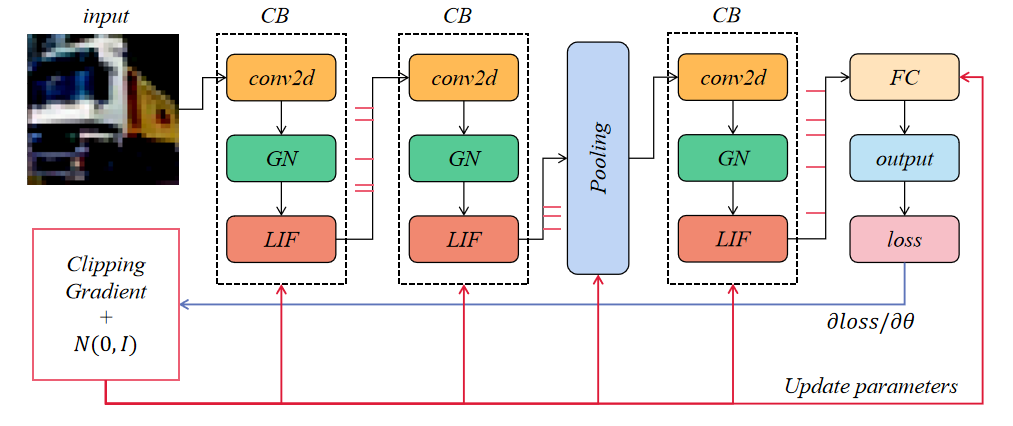}
        \end{center}
        \caption{The data pipeline of DPSNN. The forward process transmits information as the normal SNN. The backward process clips the gradient and add noise to the clipped gradient (gradient perturbation).
        CB means convolution block. The Pooling module can be different pooling operations, including max pooling(MaxPool), average pooling(AvgPool), and the temporal enhanced pooling(TEP), which we will introduce in the next section.}\label{DPSNN}
    \end{figure}

    Here, we propose DPSNN. DPSNN uses LIF neurons as Eq.~\ref{lif}$\sim $\ref{eq3}. 
    As shown in Fig.~\ref{DPSNN}, the network structure consists of several basic blocks: convolution block(CB), Pooling and fully-connected layer(FC).
    We denote the input of CB at time $t$ 
    as $X^{t}$, CB can be described as:
    \begin{equation}
        I^t=GN(conv(X^t))
        \label{conv}
    \end{equation}
    \begin{equation}
        V^t=\lambda V^{t-1}(1-O^{t-1})+(1-\lambda )I^t
        \label{CB}
    \end{equation}
    \begin{equation}
        O^t=g(V^t)
        \label{Fire}
    \end{equation}

    Here $V^t$, $I^t$, and $O^t$ mean the membrane potential, input current and output spikes of all LIF neurons in CB. 
    $GN$ means group normalization (GN)~\citep{wu2018group} and $conv$ means convolution layer. $\lambda $ and $g$ are the same as we described above
    . The Pooling module can be different pooling operations, including max pooling (MaxPool), average pooling (AvgPool), and the TEP, which we will introduce in the next section.
    For the output layer, we used the accumulated membrane potential as the final output as used in~\citep{kim2020revisiting}. The network propagates the information in the forward process as the normal SNN. 
    
    The loss function is set with cross-entropy loss. 
\begin{equation}
        L=-\sum_{i = 1}^{l(d)} y_i\log (\frac{exp(V_{i}^{T,d}/T)}{\sum_{j = 1}^{l(d)} exp(V_{i}^{T,d}/T) } )
        \label{loss}
    \end{equation}

       $y_i$ is the true label. For the backward process of the network, we first calculated the gradient of each parameter. The details are shown below.
    \begin{equation}\label{dloss}
        \begin{aligned}
            \frac{ \partial L }{ \partial w_{ij}^{n} }&=
            \sum_{t = 1}^{T}\frac{ \partial L }{ \partial V_i^{t,n} }
            \frac{ \partial V_i^{t,n} }{ \partial w_{ij}^{n} }
        \end{aligned}
    \end{equation}

    We used the surrogate gradient mentioned above combined with the backpropagation through time (BPTT) to get the gradient of the corresponding time and layer. 
   
    \begin{equation}\label{BP}
        \begin{aligned}
            \frac{ \partial L }{ \partial V_i^{t,n} }&=
            \sum_{j = 1}^{l(n+1)}\frac{ \partial L }{ \partial V_i^{t,n+1} }
            \frac{ \partial V_i^{t,n+1}}{ \partial V_i^{t,n}}
            +\frac{ \partial L}{ \partial V_i^{t+1,n}}
            \frac{ \partial V_i^{t+1,n}}{ \partial V_i^{t,n}}\\
            &=\sum_{j = 1}^{l(n+1)}\frac{ \partial L }{ \partial V_i^{t,n+1} }
            \frac{ \partial V_i^{t,n+1}}{ \partial o_i^{t,n}}
            \frac{ \partial g}{ \partial V_i^{t,n}}
            +\frac{ \partial L}{ \partial V_i^{t+1,n}}
            (1-o_i^{t,n})
        \end{aligned}
    \end{equation}
    
    We can get the gradients of parameters $\frac{ \partial L }{ \partial \theta _m}$
    where $\theta _m$ means the vector of all the $w_{ij}^{n}$ after $m$
    times parameters update. To restrict the influence of each sample to the gradient, we clipped the gradient with a certain range with $l_2$ norm. 
    \begin{equation}
        \bar{v}_k=\frac{ \partial L }{ \partial \theta _m}/\max 
            \left\{1, \left\lVert \frac{ \partial L }{ \partial \theta _m}\right\rVert _{2} /R \right\} 
        \label{clip}
    \end{equation}
    
    Then the algorithm apply Gaussian mechanism controlled by the noise scale
    $\sigma $ to clipped gradients, and updated the weights of the network 
    according to the optimization method such as SGD or Adam.
    \begin{equation}
        u_m = \sum_{k = 1}^{B} \bar{v}_k+\sigma R\cdot N(0,I) 
        \label{noise}
    \end{equation}
    \begin{equation}
        \theta _{m+1}=Optim(\theta _{m}, u_m)
        \label{update}
    \end{equation}
    Where $B$ is the mini-batch size and $Optim$ is the optimization method. As the gradients have been clipped,
    it is obviously that the global sensitivity of the function $\sum_{k = 1}^{B} \bar{v}_k$ is bounded by $R$. 
    According to Theorem \ref{gm}, the privacy bound of Eq.~\ref{noise} is $(\alpha ,\alpha /(2\sigma ^2))$-RDP.
    DPSNN iteratively optimizes the initial parameters $\theta _0$ to $\theta _E$, which is a composition of Gaussian mechanisms. 
    $E$ is the number of iterations. The RDP privacy bound after each 
    iteration can be calculated and transformed to $(\varepsilon ,\delta )$-DP by the Opacus framework~\citep{opacus}. The privacy guarantee will decrease after every
    iteration, which means the privacy bound $\varepsilon $ increases over iteration times. The complete algorithm is shown in Algorithm.\ref{alg1}.

    \begin{algorithm}[htb]
    \caption{The forward propagation, backpropagation and parameters updation of DPSNN}
    \label{alg1} 
    \begin{algorithmic}
    \REQUIRE ~~\\ 
        Mini-batch $X=\{x_1,...,x_B\}$, network model $N$ has $d$ layers. \\
        Time window $T$, loss function $L$, parameters $\theta _m$, learning rate $\eta _t$. \\
        noise scale $\sigma $, gradient norm bound $R$.
    \ENSURE ~~\\ 
    \For{$k=1$ to $B$}{
    \STATE Initialize all the membrane potential and spikes as zero
    
    \For{$t=1$ to $T$}{
        \STATE Forward propagation $N(x_k)$ 
    }
    \STATE Compute the cross-entropy loss as Eq.\ref{loss}
    \STATE Compute the $\frac{ \partial L }{ \partial V^{t,d} } $ for $t=1$ to $T$.

    \For{$t=T$ to 1}{
        \For{$n=d-1$ to 1}{
            \STATE Backpropagation as Eq.\ref{BP}
        }
    }
    \STATE Compute the gradient $\frac{ \partial L }{ \partial \theta _m} $ according to Eq.\ref{dloss}
    \STATE Clip the gradient as Eq.\ref{clip}
    }
    \STATE Integrate the gradients and add the noise as Eq.\ref{noise}
    \STATE Update parameters as Eq.\ref{update}
    \STATE Calculate the privacy bound $\varepsilon $
    \RETURN $\theta _{m+1}$;
    \end{algorithmic}
\end{algorithm}

\subsection{Temporal Enhanced Pooling}\label{TEP_section}

As introduced above, DP injects noise into the gradient, and more noise means stricter privacy protection, which inevitably causes performance degradation. Most of the traditional SNN training is based on  MaxPool or AvgPool. We first train DPSNN on CIFAR10 and CIFAR10-DVS with different pooling methods and pooling layer kernel sizes to verify the importance of pooling layers.
The privacy bound for the two datasets is $\varepsilon =8$, $\delta =1e-5$. 

    \begin{figure}[h!]
        \begin{center}
        \includegraphics[width=0.9\linewidth]{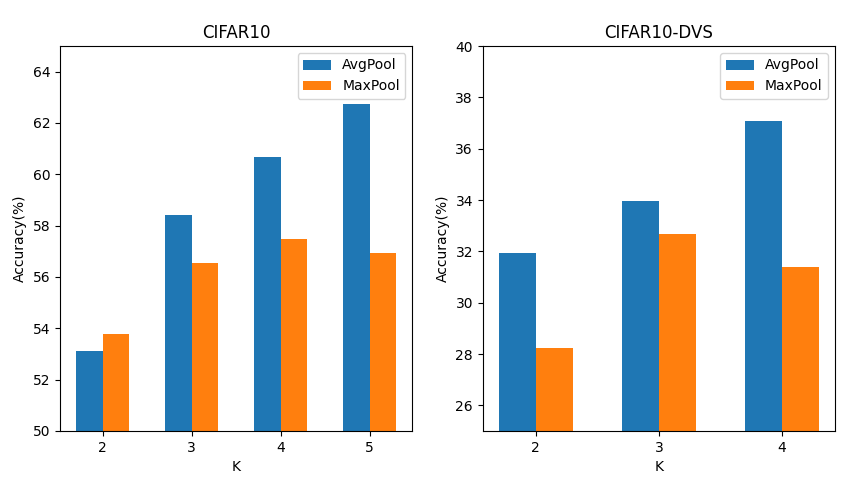}
        \end{center}
        \caption{This figure shows the test accuracy of DPSNN using different pooling methods and kernel sizes for the pooling layer. The parameter $K$ denotes the kernel size, and the strides for all pooling operations are set to 2.}\label{pool_k}
    \end{figure}

 As shown in Fig.\ref{pool_k}, different pooling settings have a significant influence on performance. MaxPool outputs the maximum spike in the neighborhood, while the maximum spike value in SNN is 1, which undoubtedly causes significant information loss. On the other hand, AvgPool outputs the average spike activity in the neighborhood without reasonably assigning different importance to different positions. In addition to spatial information transmission through layers, spiking neurons accumulate membrane potential over time and fire spikes, which contain rich temporal information. Considering that neurons with higher fire rates have a more significant positive effect on information transmission, we propose the TEP. 

    \begin{figure}
        \begin{center}
        \includegraphics[width=0.9\linewidth]{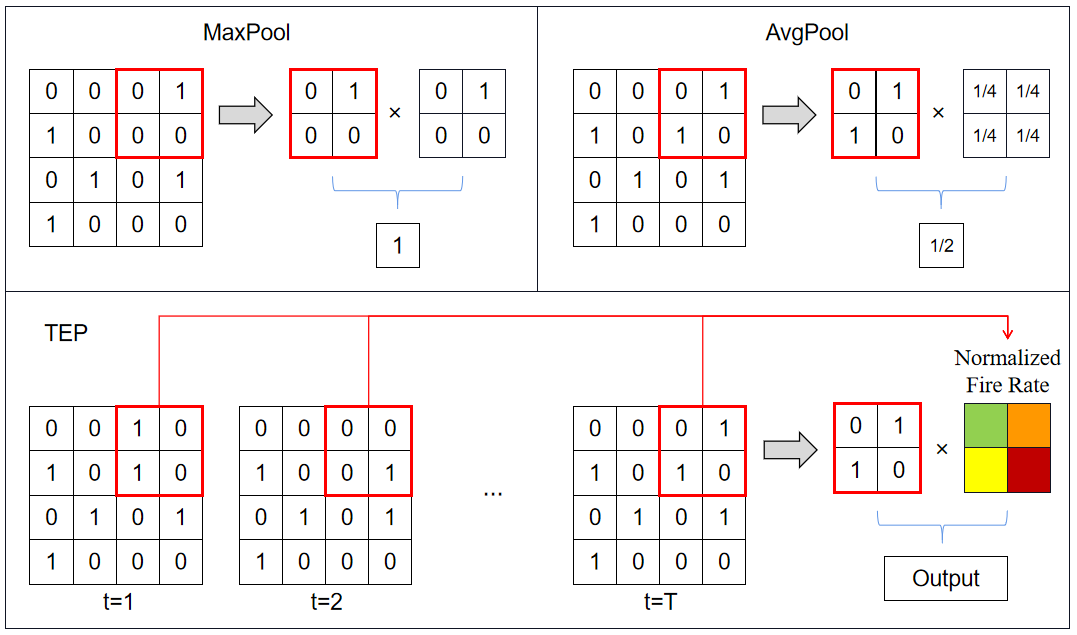}
        \end{center}
        \caption{The Illustration of MaxPool, AvgPool and TEP. As shown in the figure, TEP assigns weights to each position of the feature map based on the fire rate, utilizing the temporal information of neurons.}\label{TEP}
    \end{figure}

As shown in Fig.\ref{TEP}, TEP assigns higher weights to the neurons with higher fire rates, utilizing the temporal information of neurons to enable the more active neurons to contribute more to information transmission. 
We denote the feature map before the TEP module at time $t$ as $O^t$, and the TEP can be 
described as follows:
    \begin{equation}
        F= \sum_{t = 1}^{T} O^{t} /T
        \label{fire_rate}
    \end{equation}
    \begin{equation}
        \hat{F} = IN(F)
        \label{normalization}
    \end{equation}
    \begin{equation}
        X^t=AvgPool(O^t (\hat{F} + 1))
        \label{time}
    \end{equation}

    Eq.\ref{fire_rate} calculates the fire rate of each neuron in the feature map, which contains 
    temporal information of SNN. $IN$ in Eq.\ref{normalization} is instance normalization (IN) \citep{ulyanov2016instance}.
    We use normalization here because the spikes may be sparse in SNN, which means the fire rate will be too low. Moreover, \citet{davody2020effect} shows that the normalization method can strengthen the network's robustness to noise and positively impact private training.
    Eq.\ref{time} combines temporal features with spatial features and gets the output of 
    TEP $X^t$.
    
\section{Experiments}\label{experiment}
In this section, we show the performance of DPSNN with TEP and conduct ablation experiments to verify the superiority of TEP. All the experiments are conducted 
with Nvidia Titan RTX GPU. The implementation of SNN models are based on BrainCog framework~\citep{zeng2022braincog}, and the code for DPSNN is in the folder $/examples/Snn\_safety/DPSNN$ of the BrainCog repository on github (\href{https://github.com/BrainCog-X/Brain-Cog}{https://github.com/BrainCog-X/Brain-Cog}). The time steps $T=10$, the leak rate $\lambda =0.5$, the threshold $V_{th}=0.5$ 
and the optimizer is AdamW and $\delta =1e-5$ for
all datasets in this section. Due to the large noise impact in DP training, all DP models are 
trained five times, and their test accuracy mean and variance are calculated.

\subsection{Datasets and Network Structures}

We test the performance of DPSNN using both static datasets and neuromorphic datasets. The static dataset include CIFAR10, MNIST\citep{lecun1998mnist} and Fashion-MNIST\citep{xiao2017fashion}. CIFAR10 is a 10-classes dataset that contains 50000 labeled training
samples and 10000 labeled testing samples. Each sample is 3-channel 32$\times $32 image sourced from
real world. MNIST and Fashion-MNIST are both 10-classes datasets consisting of 60000 training samples and 10000 testing samples, and each sample is 28$\times $28 grayscale image. The neuromorphic dataset include N-MNIST\citep{orchard2015converting} and CIFAR10-DVS\citep{li2017cifar10}. The N-MNIST dataset converts the 
MNIST dataset into a spiking version. It consists of 60000 training samples and 10000 testing samples as MNIST; each sample is 28$\times $28 pixels. The CIFAR10-DVS dataset is a spiking version of CIFAR10. It converts 10000 static images from CIFAR10 into 10000 event streams by dynamic vision sensor. We choose 9000 samples for training and 1000 samples for testing in our experiments and resize the samples to 48$\times $48 pixels using interpolation.

    \begin{table}
        \centering
        \caption{The network structure for three datasets. CB(64,3,1) denotes a convolution block which has 64 output channels, and the kernel
        size and padding of the convolution layer are 3 and 1. TEP($K$) denotes the kernel size of the TEP operation is $K$, and all the strides in the TEP operations in our experiments are set to 2.
        FC(10) denotes fully-connected layer which has 10 output channels.}
        \label{structure}
        \begin{tabular}{cccc}
            \hline
           \multirow{3}{*}{Datasets} & MNIST &  &  \\
            & Fashion-MNIST & CIFAR10 & CIFAR10-DVS \\
            & N-MNIST &  &  \\
            \hline
           \multirow{11}{*}{Structure} & CB(32,7,0) & CB(64,3,1)$\times $2 & CB(64,3,1)$\times $2 \\
           \cline{2-4}
            &TEP(2) & TEP(5) & TEP(4)\\
            \cline{2-4}
            &CB(64,4,0) & CB(128,3,1)$\times $2 & CB(128,3,1) \\
            \cline{2-4}
            &TEP(2) & TEP(5) &TEP(4) \\
            \cline{2-4}
            &FC(10) & CB(128,3,1)$\times $2 & CB(256,3,1)$\times $2\\
            \cline{2-4}
            & & TEP(5) &  TEP(4)\\
            \cline{2-4}
            & & CB(256,3,1)$\times $2 &CB(512,3,1) \\
            \cline{2-4}
            & & Global Pooling & TEP(4)\\
            \cline{2-4}
            & & FC(10) & CB(1024,3,1)$\times $2\\
            \cline{2-4}
            & &  & Global Pooling\\
            \cline{2-4}
            & &  & FC(10) \\
            \hline
        \end{tabular}
    \end{table}

    We use different network structures for different datasets, as shown in Tab.\ref{structure}. A relatively small-scale network trains the MNIST, Fashion-MNIST and N-MNIST datasets. We use the VGG structure to train CIFAR10 and
    CIFAR10-DVS. CB(64,3,1) denotes a convolution block that has 64 output channels, and the kernel
    size and padding of the convolution layer are 3 and 1. TEP($K$) denotes the kernel size of the TEP is $K$, and all the strides in the TEP in our experiments are set to 2.
    FC(10) denotes a fully-connected layer that has 10 output channels. We use GN in the CB block, and the
    number of groups is 16.
    
\subsection{Static Datasets}
    
For the CIFAR10 dataset, we train the DPSNN to $\varepsilon =8$ in 80 epochs,
the gradient norm bound $R=6$, the batch size $B=1024$, and the learning rate is set to 0.001. For MNIST and Fashion-MNIST dataset, we train the DPSNN to $\varepsilon =3$ in 20 epochs,
the gradient norm bound $R=2$, the batch size $B=1024$, and the learning rate is set to 0.005.
 The DPSNN can achieve 97.71\% mean test accuracy on MNIST, 85.72\% on Fashion-MNIST, and 65.70\% on CIFAR10.

\begin{figure}[h!]
        \begin{center}
        \includegraphics[width=0.9\linewidth]{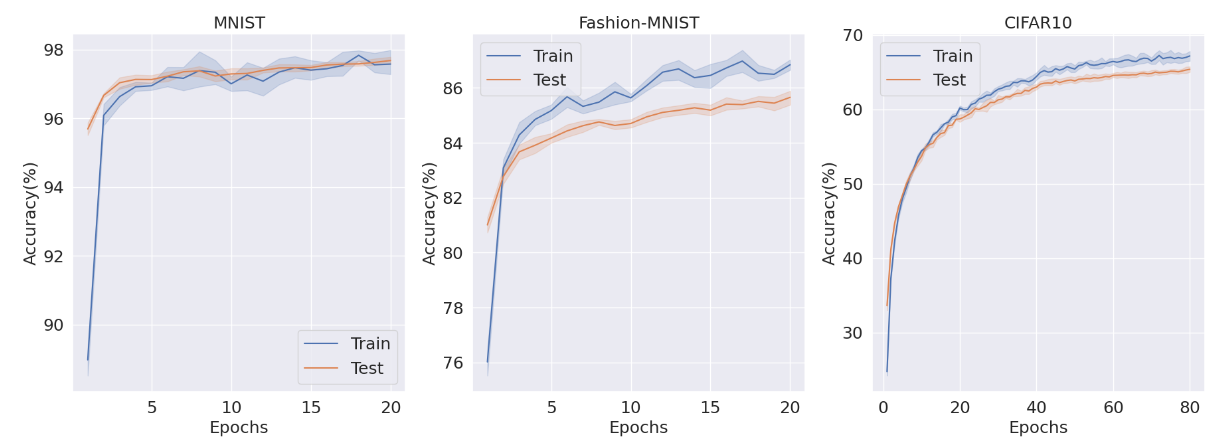}
        \end{center}
        \caption{The curve of the mean test accuracy and training accuracy of DPSNN on static datasets over training epochs, 
        with the shaded area representing the variance.}\label{TEP_static_result}
    \end{figure}

As shown in Fig.\ref{TEP_static_result}, our DPSNN achieves a favorable trade-off between privacy and performance.
For example, when training on the CIFAR10 dataset, stopping at 40 epochs just results in a slight mean test accuracy reduction to 64.06\%. At the same time, the privacy bound $\varepsilon$ can be reduced from 8 to 5.47, which means the privacy guarantee becomes better.

\subsection{Neuromorphic Datasets}

\begin{figure}[h!]
        \begin{center}
        \includegraphics[width=0.9\linewidth]{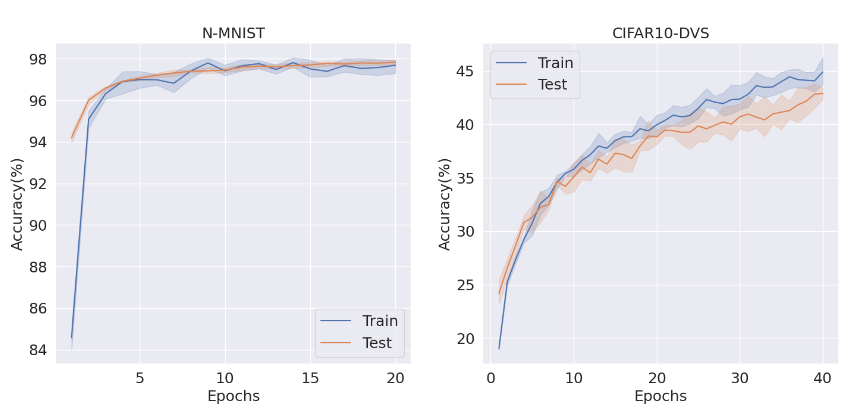}
        \end{center}
        \caption{The curve of the mean test accuracy and training accuracy of DPSNN on neuromorphic datasets over training epochs, 
        with the shaded area representing the variance.}\label{TEP_neuromorphic_result}
    \end{figure}

For the N-MNIST dataset, we adopt the same training settings as those for MNIST and Fashion-MNIST. On CIFAR10-DVS, the batch size $B=512$, and the resting training settings are the same as those for the CIFAR10. The performance of DPSNN with TEP on the neuromorphic datasets are shown in Fig.\ref{TEP_neuromorphic_result}. The mean test accuracy of DPSNN  can reach 43.24\% on CIFAR10-DVS and 97.78\% on N-MNIST.

\section{Discussion}\label{s}

Based on the CIFAR10 and CIFAR10-DVS datasets, in this section, we first perform the ablation experiments to analyze the effect of TEP. Then we analyze the impact of different neuron types on the privacy protection of SNN.

\subsection{Ablation Experiments}

We compare the performance of MaxPool, AvgPool, and TEP with the same kernel size to show the superiority of TEP. The training settings are the same as mentioned above. 

\begin{figure}[h!]
        \begin{center}
        \includegraphics[width=0.9\linewidth]{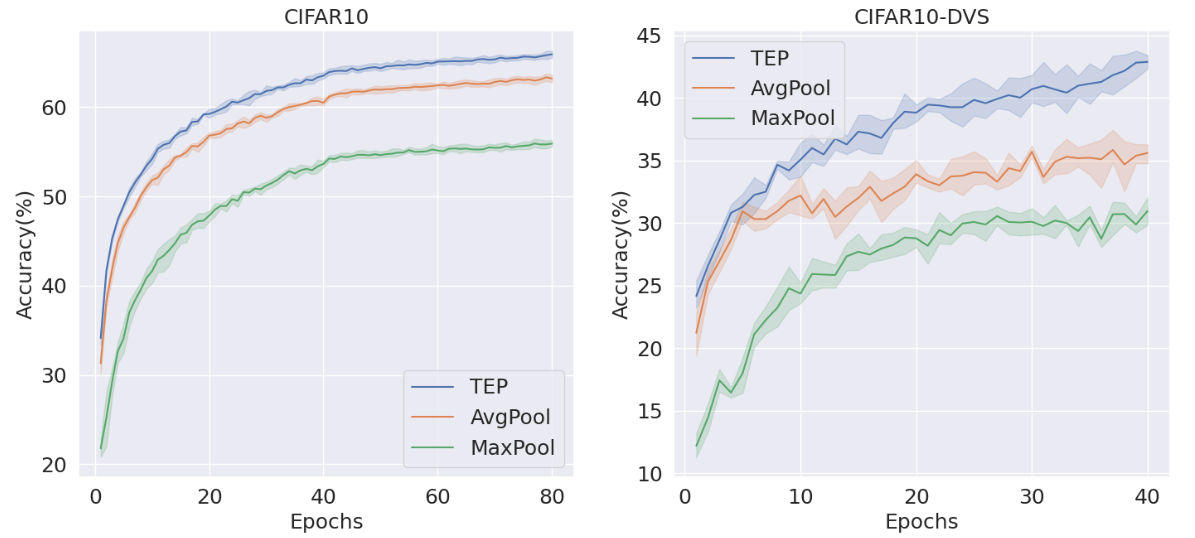}
        \end{center}
        \caption{The curve of the mean test accuracy of DPSNN using different pooling operations, with the shaded area representing the variance.}\label{TEP_AP_MP}
    \end{figure}

Fig.\ref{TEP_AP_MP} displays the curve of the mean testing accuracy of DPSNN with MaxPool, AvgPool, and TEP, with the shaded area indicating the variance.
The figure shows that networks utilizing TEP demonstrate a faster convergence rate and achieve higher accuracy than those using other pooling operations. This suggests that using TEP can improve performance under different levels of privacy guarantee. Therefore, the use of TEP in DPSNN can be considered 
as an effective method for improving both the convergence rate and accuracy of the network, making it 
suitable for various applications that require privacy protection.

    \subsection{The Influence of Different Neuron Types}\label{lif_if}
\begin{figure}[h!]
        \begin{center}
        \includegraphics[width=0.9\linewidth]{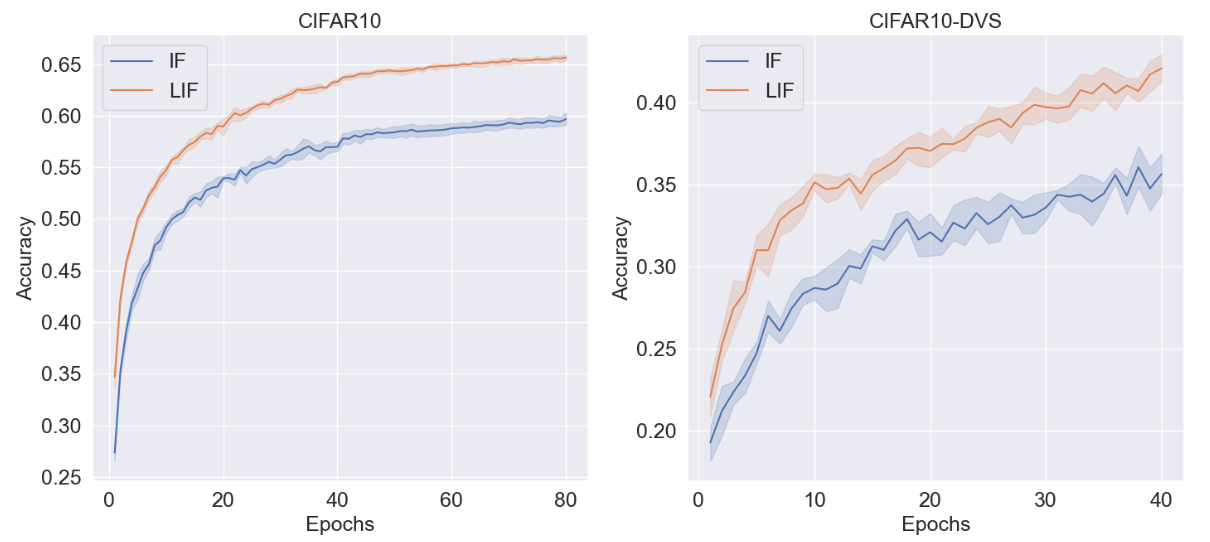}
        \end{center}
        \caption{The performance of the DPSNN using 
        LIF neurons and IF neurons.}
        \label{IFLIF}
    \end{figure}

    In addition to the LIF neurons, Integrate-And-Fire (IF) neuron is another commonly used neuron model 
    in the deep SNN. The dynamics of IF neurons can be described as follows:
    \begin{equation}\label{eq1}
        C \frac{dV(t)}{dt} = I(t)
    \end{equation}
    where $C $ is the capacitance of the cell membrane. We can derive the iterative IF model as follows:
    \begin{equation}\label{eq2}
        V_{i}^{t+1,n}=V_{i}^{t,n}(1-o_{i}^{t,n})+
        \sum_{j = 1}^{l(n-1)}w_{ij}^{n}o_{j}^{t+1,n-1}
    \end{equation}
    We use IF neurons instead of LIF neurons to train DPSNN. Fig.\ref{IFLIF} shows the comparison of the 
    experimental results obtained using two types of neurons.
    
    As shown in the Fig.~\ref{IFLIF}, LIF neurons exhibit superior performance over IF neurons on both CIFAR10 as well as CIFAR10-DVS datasets with the same privacy-preserving bounds.
    The leak rate enables the network to pay attention to different information influences at different time steps. 
    The farther the time steps, the smaller the influence. 
    The impact of this temporal information increases the feature extraction ability of TEP and
    consequently improves the performance of DPSNN.

    \subsection{Conclusion and future work}
    This study combines DP algorithm with SNN for the first time, providing strong privacy protection for SNN. At the same time, to avoid the performance degradation caused by the noise injection of DP, we design the TEP operation. This operation balances performance and privacy protection for the SNN by incorporating the temporal information of spiking neurons into the pooling operation. We conduct experiments on both static datasets and neuromorphic datasets, and the experimental results show that our algorithm can maintain high accuracy while providing strong privacy protection.
    
    DPSNN also has some limitations. The implementation of large-scale DPSNN is still challenging, because the spike distribution 
    in the deep layers of the network can easily become either too sparse or too dense. 
    As the experiments in section~\ref{lif_if} have shown that LIF neuron perform better than IF neuron significantly,
    we can design spiking neuron model with more diverse temporal dynamics in the future.

    \section{Acknowledgements}

    This work was supported by the Strategic Priority Research Program of Chinese Academy of Sciences (XDB32070100); the Chinese Academy of Sciences Foundation Frontier Scientific Research Program (ZDBS-LY- JSC013).

\bibliography{test}
\bibliographystyle{elsarticle-harv}



\end{document}